\pdfoutput=1

\documentclass[11pt]{article}

\usepackage[]{EMNLP2023}

\usepackage{times}
\usepackage{latexsym}
\usepackage{booktabs}
\usepackage{makecell}
\usepackage{adjustbox}
\usepackage{multirow}
\usepackage{amsmath}
\usepackage{amssymb}
\usepackage{enumitem}

\usepackage[T1]{fontenc}

\usepackage[utf8]{inputenc}

\usepackage{microtype}

\usepackage{inconsolata}

\title{Prompting and Evaluating Large Language Models for Proactive Dialogues: Clarification, Target-guided, and Non-collaboration}

\author{Yang Deng$^{1}$, Lizi Liao$^2$, Liang Chen$^{3}$, Hongru Wang$^{3}$,  Wenqiang Lei$^{4}$, Tat-Seng Chua$^1$ \\
  $^1$National University of Singapore 
  $^2$Singapore Management University\\
  $^3$The Chinese University of Hong Kong 
  $^4$Sichuan University\\
  \texttt{\{ydeng,dcscts\}@nus.edu.sg},
  \texttt{lzliao@smu.edu.sg}\\
  \texttt{\{lchen,hrwang\}@se.cuhk.edu.hk},
 \texttt{wenqianglei@gmail.com}
  \\}

\begin{document}
\maketitle
\begin{abstract}
Conversational systems based on Large Language Models (LLMs), such as ChatGPT, show exceptional proficiency in context understanding and response generation. However, they still possess limitations, such as failing to ask clarifying questions to ambiguous queries or refuse users' unreasonable requests, both of which are considered as key aspects of a conversational agent's proactivity. This raises the question of whether LLM-based conversational systems are equipped to handle proactive dialogue problems. In this work, we conduct a comprehensive analysis of LLM-based conversational systems, specifically focusing on three key aspects of proactive dialogues: clarification, target-guided, and non-collaborative dialogues. To trigger the proactivity of LLMs, we propose the Proactive Chain-of-Thought prompting scheme, which augments LLMs with the goal planning capability over descriptive reasoning chains. Empirical findings are discussed to promote  future studies on LLM-based proactive dialogue systems. 
\end{abstract}

\section{Introduction}
Conversational systems are envisioned to provide social support or functional service to human users via natural language interactions. 
Most research typically centers around a system's response capabilities, such as understanding the dialogue context~\cite{tod-bert,sigdial22-unidu,www22-use} and generating appropriate responses~\cite{dialogpt,blenderbot}. 
The popularity of conversational systems has grown unprecedentedly with the advent of ChatGPT, which showcases exceptional capabilities of context understanding and response generation with large language models (LLMs). 
Recent studies observe that, compared with current fine-tuned state-of-the-art (SOTA) methods, ChatGPT can still achieve competitive performance under zero-shot setting on different dialogue problems, such as the knowledge-grounded dialogues~\cite{mmmeval-chatgpt}, task-oriented dialogues~\cite{zhang2023sgptod}, and emotion-aware dialogues~\cite{chatgpt-emo-dial}.

Despite the strength of ChatGPT, there are still several limitations\footnote{as stated in  \url{https://openai.com/blog/chatgpt/}.}, such as failing to ask clarification questions to ambiguous user queries or refuse problematic user requests. These kinds of capabilities are typically regarded as the \textit{proactivity} of the conversational system~\cite{proactive-survey}, where the system can create or control the conversation to achieve the conversational goals by taking initiative and anticipating impacts on themselves or the human users. Thus, it raises the question: \textit{Are these LLM-based conversational systems equipped to manage proactive dialogue problems?} 

In this work, we conduct the first comprehensive analysis of LLM-based conversational systems on three common aspects of proactive dialogues, including 1) clarification in information-seeking dialogues~\cite{abg-coqa,pacific} where the system is required to proactively ask clarification questions when encountering ambiguity in user queries; 2) target-guided open-domain dialogues~\cite{acl19-tgoc,acl19-proactive} where the system is required to proactively lead the conversation towards the designated target; and 3) non-collaborative task-oriented dialogues~\cite{aaai20-non-collab,iclr20-non-collab,acl23-tutorial} where the system and the user do not share the same conversational goal while the system aims to strategically reach a consensus with the user.

\begin{figure*}
\setlength{\abovecaptionskip}{5pt}   
\setlength{\belowcaptionskip}{5pt}
\centering
\includegraphics[width=\textwidth]{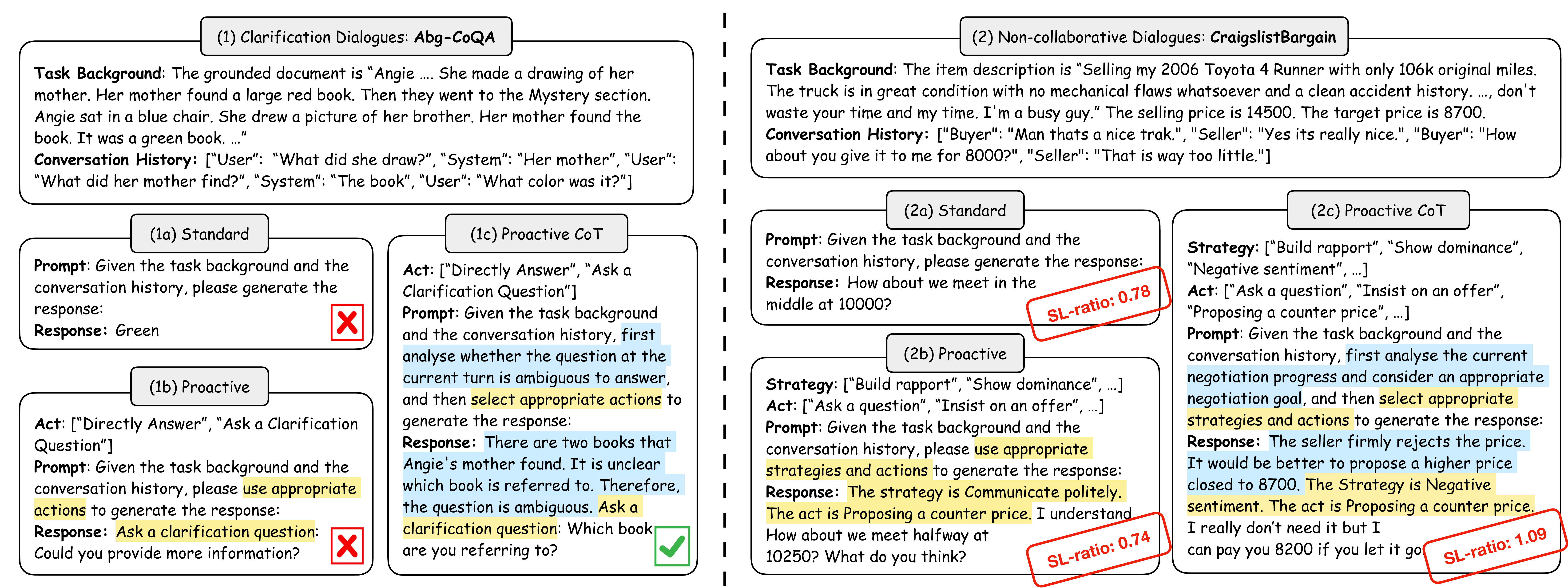}
\caption{Examples of three kinds of prompting schemes for proactive dialogues. In the example of non-collaborative dialogue, the system plays the role of "Buyer", and the sale-to-list (SL) ratio shows the effectiveness of negotiation, which is calculated by ($\text{listed price} - \text{bargain price})/(\text{listed price} - \text{buyer target price})$. The higher ratio means the current bargain price is closer to the target.}
\label{example}
\vspace{-0.5cm}
\end{figure*}

Motivated by the emergent capabilities of LLMs \cite{emergent_ability,cot} on reasoning over texts, some recent studies investigate in-context learning or chain-of-thought prompting schemes on planning \cite{monologue} or taking actions~\cite{react} in interactive environments. 
Similarly, strategy learning and goal planning attach great importance in proactive dialogue systems. In order to enhance the proactivity of LLM-based conversational systems, we design the proactive chain-of-thought prompting (ProCoT) scheme. 
As shown in Figure~\ref{example}, with standard prompting, LLM-based systems directly provide a randomly-guessed answer to the ambiguous user question (1a), or generate a general bargain response without any negotiation strategy (2a). 
When providing the system with options to take different dialogue acts (proactive prompting), the generated responses are unaware of the conversational goal, such as generating under-specified clarification questions (1b) and conservative negotiation responses (2b). 
To this end, ProCoT first instructs the system to generate descriptive thoughts about intermediate steps of reasoning and planning for reaching the conversational goal, and then make the decision of the next action to take. Finally, the system generates an appropriate response based on the decided action (1c \& 2c).

We conduct extensive experiments with two LLM-based conversational systems, including ChatGPT and an open-sourced model, Vicuna~\cite{vicuna}. 
With the aforementioned three types of prompting schemes, we compare these LLM-based conversational systems with fine-tuned SOTA dialogue models. 
The main contributions of this work can be summarized as follows:
\begin{itemize}[leftmargin=*]
    \item This work presents the first comprehensive evaluation on the proactivity of LLM-based dialogue systems, including the handling of clarification, target-guided, and non-collaborative dialogues. 
    \item We design the proactive chain-of-thought prompting scheme to endow LLM-based dialogue systems with the capability of planning and taking the initiative towards the conversational goal. 
    \item Specifically, the main findings of the evaluation of LLM-based dialogue systems include: 1) They barely ask clarification questions when encountering ambiguous queries, and ProCoT largely overcomes this issue, though the performance is still unsatisfactory in domain-specific applications ($\S \ref{sec:clari}$). 2) They are proficient at performing topic shifting towards the designated target, but tend to make aggressive topic transition. ProCoT further improves this capability by planning a smoother transition ($\S \ref{sec:target}$).  3) They fail to make strategic decision and tend to compromise with the opponent. The key challenge is how to effectively optimize the strategy learning ($\S \ref{sec:non-colla}$). 
\end{itemize}

\section{Related Works}
\paragraph{Proactive Dialogues.}
Recent years have witnessed many advanced designs on developing proactive dialogue systems~\cite{wsdm23-proactive} for various applications. For example, target-guided dialogues aim to proactively lead the conversation to either a designated target topic~\cite{acl19-tgoc} or a pre-defined knowledge entity~\cite{acl19-proactive}. Existing studies typically adopt keyword transition~\cite{aaai20-tgoc,aaai21-tgoc} or knowledge graph reasoning~\cite{coling22-topkg,sigir22-proactive} techniques to proactively plan the topic thread towards the target. 
Besides, in information-seeking dialogues, proactive dialogue systems can ask clarification questions for clarifying the ambiguity of the query or question in conversational search~\cite{emnlp21-clariq} and question answering~\cite{abg-coqa,pacific}. 
In addition, under the non-collaborative setting, the system and the user have competing goals towards the task completion but the system aims to proactively reach an agreement favorable to itself~\cite{iclr20-non-collab}, such as negotiating a product price~\cite{emnlp18-negotiate} or persuading users to make a donation~\cite{acl19-persuasion}. 

\paragraph{Large Language Models for Dialogues.}
Previous dialogue systems, such as DialoGPT~\cite{dialogpt}, Meena~\cite{meena}, BlenderBot~\cite{blenderbot}, LaMDA~\cite{lamda}, typically fine-tune pre-trained language models on public dialogue data. Inspired by the success of ChatGPT, recent practices build dialogue systems through conducting supervised fine-tuning on open-source large language models, such as LLaMA~\cite{llama}, with either constructed instruction-following examples (\textit{e.g.}, Alpaca~\cite{alpaca}) or distilled conversation data (\textit{e.g.}, Vicuna~\cite{vicuna}) from ChatGPT. 
As all these LLM-based dialogue systems are trained to follow the user's instruction, it remains a question on whether these systems can take the initiative for handling proactive dialogues.

\paragraph{Prompting in Dialogue Systems.}
To induce knowledge from LLMs, various prompting methods are designed for zero-shot or few-shot learning in dialogue applications, such as task-oriented dialogues~\cite{emnlp21-prompt-tod,aaai22-prompt-tod}, knowledge-grounded dialogues~\cite{emnlp22-findings-prompt-kgc,acl22-findings-prompt-kgc,acl23-knowdial}, and open-domain dialogues~\cite{eacl23-findings-prompt-socialdial,acl23-prompt-ood,cue}. For example, \citet{acl23-prompt-mixed} propose to prompt LLMs for controllable response generation in emotional support and persuasion dialogues, conditioned on the ground-truth dialogue strategies. 
In this work, we aim at prompting LLMs to proactively interact with the users.

\section{Prompting LLMs to be Proactive}\label{sec:prompt}
As presented in Figure~\ref{example}, we describe the prompting schemes, including the standard, proactive, and proactive chain-of-thought (ProCoT) prompting.  
\paragraph{Standard Prompting.}
In order to instruct LLMs to perform specific dialogue tasks, the typical prompting scheme can be formulated as 
\begin{equation}\small
    p(r|\mathcal{D},\mathcal{C}). 
\end{equation}
Given the task background $\mathcal{D}$ and the conversation history $\mathcal{C}$, instruct the LLM to generate the response $r$. 
In specific, the task background can be the grounded document in clarification dialogues or the target description in target-guided dialogues. 

\paragraph{Proactive Prompting.}
Proactive prompting aims to provide alternative options for LLMs to decide what kinds of actions should be taken in the response, instead of simply responding to the instruction. 
It can be formulated as:
\begin{equation}\small
p(a,r|\mathcal{D},\mathcal{C},\mathcal{A}) . 
\end{equation}
Given the task background $\mathcal{D}$, the conversation history $\mathcal{C}$, and a set of possible dialogue acts $\mathcal{A}$, instruct the LLM to select the most appropriate dialogue act $a\in\mathcal{A}$ and then generate the response $r$.  
For example, the dialogue act can be \textit{Ask a Clarification Question} or \textit{Directly Answer the Question} in clarification dialogues, different negotiation strategies in non-collaborative dialogues, or different conversation topics in target-guided dialogues. 

\paragraph{Proactive Chain-of-Thought Prompting.}
In order to endow LLMs with the capability of planning and taking the initiative towards the ultimate goal, we develop the proactive chain-of-thought prompting scheme---ProCoT. It involves the analysis of the next action to take by performing dynamic reasoning and planning for reaching the conversational goal. 
ProCoT can be formulated as:
\begin{equation}\small
p(t,a,r|\mathcal{D},\mathcal{C},\mathcal{A}), 
\end{equation}
where $t$ is the thought description for the decision-making process of the next action. For example, in clarification dialogues, $t$ can be the ambiguity analysis of the current user question as in Figure~\ref{example}(1c). While in non-collaborative dialogues, $t$ can be the goal completion analysis of the current negotiation progress as in Figure~\ref{example}(2c).

\begin{table}
\setlength{\belowcaptionskip}{0pt}
    \centering
    \setlength{\tabcolsep}{1mm}{
    \begin{adjustbox}{max width=0.48\textwidth}
    \begin{tabular}{lcccccccc}
    \toprule
       &&& \multicolumn{3}{c}{Abg-CoQA} & \multicolumn{3}{c}{PACIFIC}\\
       \cmidrule(lr){4-6}\cmidrule(lr){7-9}
       &&& CNP & \multicolumn{2}{c}{CQG} & CNP & \multicolumn{2}{c}{CQG}\\
       \cmidrule(lr){4-4}\cmidrule(lr){5-6}\cmidrule(lr){7-7}\cmidrule(lr){8-9}
        Method & Shot& Prompt   & F1 & BLEU-1 & Help.  & F1 & ROUGE-2 & Help.\\
        \midrule
        Baseline & - & - & 22.1 & 36.5& 30.0 & 79.0 & 69.2 & 38.2\\
        SOTA & - & - & \underline{23.6} & \underline{38.2} & \underline{56.0} &\underline{86.9} & \underline{90.7} & \underline{80.1}\\
        \midrule
        \multirow{6}{*}{Vicuna-13B} & 0 & Standard &   -&11.3 & 0.0 &  - & 1.2 & 0.0\\
        & 1 & Standard &   -& 11.4 & 0.0 &   - & 2.5 & 0.0\\
        & 0 & Proactive & 4.1 & 13.2 & 0.0&  2.3 & 2.3 &  0.0 \\
        & 1 & Proactive &  12.1 & 13.2 & 4.5& 0.0 & 3.3 &0.0 \\
        & 0 & ProCoT &  1.4 & 21.3 & 9.1&  9.7 & 3.8 & 10.5\\
        & 1 & ProCoT &  \textbf{18.3} & \textbf{23.7} & \textbf{22.7}& \textbf{27.0} & \textbf{41.3} & \textbf{33.1}\\
        \midrule
        \multirow{6}{*}{ChatGPT} & 0 & Standard &   -&12.1 & 0.0 &   -& 2.2 & 0.0\\
        & 1 & Standard &   -&12.3 & 0.0 &   - & 2.0 & 0.0\\
        & 0 & Proactive &  22.0 & 13.7 & 17.6& 19.4 & 2.9 & 0.0\\
        & 1 & Proactive & 20.4 & \textbf{23.4}&23.5& 17.7 & 14.0 & 12.5\\
        & 0 & ProCoT &  23.8 & 21.6&32.4& \textbf{28.0} & \textbf{21.5} &26.7\\
        & 1 & ProCoT & \textbf{27.9} & 18.4&\textbf{45.9}&  27.7 & 16.2 & \textbf{35.8}\\
        
        \bottomrule
    \end{tabular}
    \end{adjustbox}}
    \caption{Experimental results on Abg-CoQA and PACIFIC datasets, whose baseline and SOTA results are adopting from \citet{abg-coqa} and \citet{pacific}. \textbf{Bold} and \underline{underlined} results denote the best performance for each LLM and the fine-tuned methods, respectively.}
    \label{tab:cnp}
\vspace{-0.5cm}
\end{table}

\section{Evaluation}
We evaluate the proactivity of LLM-based conversational systems from three perspectives, including the capability of asking clarification questions ($\S$~\ref{sec:clari}), guiding the conversation towards the designated target ($\S$~\ref{sec:target}), and strategically handling conflicting goals ($\S$~\ref{sec:non-colla}). 

\subsection{Clarification Dialogues}\label{sec:clari}
Clarification in information-seeking dialogues~\cite{cis} refers to the process of seeking further information or details to better understand the topic or question at hand. In this context, clarification is an important part of the dialogue as it helps to ensure that the information being shared is accurate and complete.

\subsubsection{Problem Definition}
Following previous studies~\cite{emnlp21-clariq,abg-coqa,pacific}, the problem of asking clarification questions can be decomposed into two subtasks: 1) \textit{Clarification Need Prediction} (\textbf{CNP}) to identify the necessity of clarification in the current turn, and 2) \textit{Clarification Question Generation} (\textbf{CQG}) to produce an appropriate clarifying question if needed. 
Given the grounded document $\mathcal{D}$ and the dialogue context $\mathcal{C}=\{q_1,a_1,...,q_{t-1},a_{t-1},q_t\}$, the dialogue system aims to first predict the binary ambiguity label $y$ on whether the current question $q_t$ needs to be clarified. If so, a corresponding clarification question should be generated as the response $a_t$ for clarifying the ambiguity. 

\subsubsection{Experimental Setups}\label{sec:exp_setup}
\paragraph{Datasets.} 
We evaluate the capability of asking clarification questions in LLM-based dialogue systems on two types of datasets: 1) \textbf{Abg-CoQA}~\cite{abg-coqa} in general domain, and 2) \textbf{PACIFIC}~\cite{pacific} in finance domain. Details on these datasets can be found in Appendix~\ref{app:dataset}. 

\paragraph{Evaluation Metrics.}
Following previous studies~\cite{abg-coqa,pacific}, we use the F1 score for the evaluation of CNP, and BLEU-1 and ROUGE-2 (F1) for the evaluation of CQG. 
In addition, since the automatic lexical matching metrics may fail to actually estimate the clarification capability of the generated clarifying questions~\cite{abg-coqa}, we also adopt human evaluation to score whether the generated question is helpful for clarifying the existing ambiguity (\textbf{Help.}). 

\paragraph{Usage of LLMs.} 
To facilitate reproducibility, we adopt a static version of ChatGPT, \textit{i.e.}, \texttt{gpt-3.5-turbo-0301}, and set the temperature to 0 for generating the deterministic outputs with the same inputs. 
In addition, we adopt an open-source LLM, \textit{i.e.}, \texttt{Vicuna-13B-delta-v1.1}
, for the evaluation. The maximum number of new tokens is set to 128 for the generation. 

\paragraph{Prompting Schemes.}
We evaluate the three prompting schemes introduced in Section~\ref{sec:prompt}, including standard, proactive, and ProCoT prompting. In addition, we report their results under both zero-shot and few-shot settings. Due to the limitation of the maximum sequence length in Vicuna (2,048 tokens), we only apply one-shot in-context learning for comparisons. 
The complete prompts adopted for evaluation is presented in Appendix~\ref{app:example}.

\subsubsection{Experimental Results}
Table~\ref{tab:cnp} summarizes the evaluation results on Abg-CoQA and PACIFIC datasets. There are several notable observations as follows:

\paragraph{LLM-based conversational systems fail to ask clarification questions.} 
Under standard prompting, both Vicuna and ChatGPT fail to ask clarification questions when encountering ambiguous queries, according to the human evaluation on the helpfulness (\textbf{Help.}) of the generated responses for clarifying ambiguity. 
Even with one-shot demonstration, in-context learning (ICL) still cannot provide them with such ability. 
Under proactive prompting, given the option of clarification, Vicuna's ability to accurately take this action is still quite limited, with the \textbf{F1} scores close to 0. In contrast, ChatGPT becomes capable of asking clarification questions on Abg-CoQA, as evidenced by the improvement on both \textbf{F1} and \textbf{Help.} scores.

\paragraph{ProCoT effectively endows LLM-based conversational systems with the capability of asking clarification questions.} 
Zero-shot ProCoT is not working in Vicuna, but one-shot ICL can largely improve the performance. 
As for Abg-CoQA, ChatGPT with zero-shot ProCoT achieves competitive performance with SOTA fine-tuned methods on the CNP task (\textbf{F1}), but the generated clarification questions are still unsatisfactory (\textbf{Help.}). 
One-shot ICL further improves the performance of ChatGPT with ProCoT to a great extent. 
The case study in Appendix~\ref{app:case-clari} shows that ProCoT also improves the explanability of asking clarification questions.

\paragraph{As for domain-specific problem, there is still a noticeable gap from the fine-tuned methods.}
Although ProCoT has already largely enhanced the capability of asking clarification questions, the performance of LLMs on the domain-specific task, \textit{i.e.}, PACIFIC (Finance), is still far behind the fine-tuned methods. 
In fact, with fine-tuning on domain-specific data, the SOTA method can achieve a remarkable performance on PACIFIC, \textit{i.e.}, 86.9 (\textbf{F1}) for CNP and 80.1 (\textbf{Help.}) for CQG, indicating the importance of domain knowledge.

\subsubsection{Error Analysis}
In order to find out the reason why LLM-based dialogue systems with ProCoT prompting fall short of handling domain-specific clarification dialogues, we randomly sample 100 error cases in clarification question generation from each dataset for analysis (all cases are generated by ChatGPT with one-shot ProCoT). We categorize these failure cases into four groups, including \textit{Wrong Aspect}, \textit{Under-specified Clarification}, \textit{Over-specified Clarification}, and \textit{Generation Error}. The details and examples can be found in the Appendix~\ref{app:error}. 
The statistics of error analysis is presented in Table~\ref{tab:error_stat}. It can be observed that the proportion of failure cases attribute to the wrong aspect and under-specified clarification in PACIFIC (Finance) is higher than that in Abg-CoQA (General). This indicates that \textbf{ChatGPT may lack of certain domain knowledge required for asking precise and specific clarification questions.}

\begin{table}[]
\setlength{\belowcaptionskip}{0pt}
    \centering
    \begin{adjustbox}{max width=0.35\textwidth}
    \begin{tabular}{lrr}
    \toprule
         & Abg-CoQA & PACIFIC \\
         \midrule
     Wrong Aspect & 21\% & \textbf{30\%}\\
     Under-spec. Clari. & 16\% & \textbf{23\%}\\
     Over-spec. Clari. & \textbf{15\%} & 5\%\\
     Generation Error & \textbf{48\%} & 42\%\\
     \bottomrule
    \end{tabular}
    \end{adjustbox}
    \caption{Statistics of error analysis.}
    \label{tab:error_stat}
    \vspace{-0.5cm}
\end{table}

\subsection{Target-guided Dialogues}\label{sec:target}
Instead of making consistent responses to the user-oriented topics, the dialogue system for target-guided dialogues is required to proactively lead the conversation topics towards a designated target~\cite{acl19-tgoc}. According to different applications, the target can be topical keywords~\cite{aaai21-tgoc}, knowledge entities~\cite{acl19-proactive}, or items to be recommended~\cite{tois23-mgcrs}.

\subsubsection{Problem Definition}
Given a target $\mathcal{D}$ that is only presented to the
agent but unknown to the user, the dialogue starts from an arbitrary initial topic, and the system needs to produce multiple turns of responses $\{u_n\}$ to lead the conversation towards the target in the end. 
The produced responses should satisfy (i) \textbf{transition smoothness}, natural and appropriate content under the given dialogue context, and (ii) \textbf{target achievement},  driving the conversation towards the designated target. 
The problem is typically decomposed into two subtasks~\cite{acl19-tgoc,aaai21-tgoc,coling22-topkg}: next topic selection and transition response generation.

\subsubsection{Experimental Setups}
\paragraph{Datasets.} 
We first conduct turn-level evaluation of the target-guided capability on a next-turn target-oriented dataset \textbf{OTTers} \cite{acl21-otters}, which requires the dialogue system to proactively bridge the current conversation topic to approach the target. Furthermore, we adopt \textbf{TGConv}~\cite{coling22-topkg} to testify the ability to guide the multi-turn conversation to the target topic as the dialogue-level evaluation. Details can be found in Appendix~\ref{app:dataset}. 

\paragraph{Automatic Evaluation Metrics.}
Following previous studies~\cite{acl21-otters,coling22-topkg}, we adopt the hits@$k$ ($k\in[1,3]$) for evaluating next topic prediction. 
Three text generation metrics, including BLEU, ROUGE-L, and METEOR scores, are used for the evaluation of response generation on the OTTers dataset. 

As for the dialogue-level evaluation on the TGConv dataset, we follow existing studies~\cite{coling22-topkg,acl23-proactive} to simulate multi-turn conversations via self-play~\cite{acl19-tgoc}, where the simulated user is unaware of the target topic. Three aspects are evaluated: 1) \textbf{Succ.} is the success rate of generating the target word within 8 turns of conversations; 2) \textbf{Turns} is the average turns of all dialogues that successfully reach the target word; and 3) \textbf{Coh.} is the contextual semantic similarity between the last utterance and the generated response, which is measured by MiniLM~\cite{minilm}. 

\paragraph{Human Evaluation Metrics.}
We also conduct the same human evaluation as \citet{coling22-topkg}, including two dialogue-level metrics with the following instructions provided for annotators: 
\begin{itemize}[leftmargin=*]
    \item Global-Coherence (G-Coh.): Whether the entire dialogue is logically and topically coherent. 
    \item Effectiveness (Effect.): How efficiently the target is achieved.
\end{itemize}
A total of 100 dialogues are generated through simulation for each method. 
Three annotators assign ratings to the generated dialogues on a scale of [0, 1, 2], where higher scores indicate better quality.

\paragraph{Baselines.}
We report the results of several fine-tuned baselines for target-guided dialogues, including GPT-2~\cite{gpt2}, DKRN~\cite{aaai20-tgoc}, CKC~\cite{aaai21-tgoc}, TopKG~\cite{coling22-topkg}, and \textsc{Color}~\cite{acl23-proactive}. 

\subsubsection{Turn-level Evaluation}\label{sec:turn}

\begin{table}[!t]
\setlength{\belowcaptionskip}{0pt}
    \centering
    \setlength{\tabcolsep}{1mm}{
    \begin{adjustbox}{max width=0.48\textwidth}
    \begin{tabular}{lccccccc}
    \toprule
       &&& \multicolumn{3}{c}{\textbf{Response Generation}} & \multicolumn{2}{c}{\textbf{Next Topic Prediction}}  \\
       \cmidrule(lr){4-6}\cmidrule(lr){7-8}
        Method & Shot& Prompt & BLEU & METEOR & R-L  & hits@1 & hits@3   \\
    \midrule
     GPT2 &- &-& 11.58& 10.26 & 17.67 & 4.39 & 15.79   \\
     DKRN&- &- & 12.86& 11.90 & 21.52 & 4.91 & 17.72  \\
     CKC &- &- & 13.34 &11.65 & 24.77& 6.87 & 21.89   \\
     TopKG &- &- &  \underline{15.35}&\underline{13.41} & \underline{27.16} & \underline{7.78} & \underline{22.06}  \\
     \midrule
     \multirow{6}{*}{Vicuna-13B} & 0 & Standard & 10.01 & 13.27 & 16.00 & 12.01 & 19.03\\
     & 1 & Standard & 10.63 & 14.81 & 17.53 & 12.10 & 16.13\\
        & 0 & Proactive & 1.41 & 18.45 & 15.45  & 9.41 & 19.89\\
        & 1 & Proactive & \textbf{13.87} & \textbf{20.96} & \textbf{21.36}  & 12.90 &\textbf{22.31}\\
        & 0 & ProCoT &5.27 & 16.59 & 15.96  & 11.56 & 18.01\\
        & 1 & ProCoT &13.38 & 19.70 & 20.62 & \textbf{15.05} & 20.70\\
     \midrule
     \multirow{6}{*}{ChatGPT} & 0 & Standard &11.34& 20.62 & \textbf{18.26}  & 13.44 & 27.69\\
     & 1 & Standard &14.41 &  19.29 & 17.73  & 15.86 & 26.34\\
        & 0 & Proactive & 14.09  &\textbf{21.06} & 15.56  & 7.53 & 22.58\\
        & 1 & Proactive & \textbf{14.74} & 19.59 & 16.29  & 8.60 & 21.23\\
        & 0 & ProCoT & 10.20& 19.57 & 15.97 & 12.63 & 23.92\\
        & 1 & ProCoT & 9.63 & 19.82 & 17.19 & \textbf{17.74} & \textbf{29.57}\\
     \bottomrule
    \end{tabular}
    \end{adjustbox}}
    \caption{Turn-level evaluation results on Next Topic Prediction and Transition Response Generation.}
    \label{tab:target_turn}
    \vspace{-0.3cm}
\end{table}

Table~\ref{tab:target_turn} shows the turn-level evaluation results on OTTers. 
There are several notable observations:

\paragraph{LLM-based dialogue systems are proficient at performing topic shifting towards the designated target.} 
According to the performance of LLMs with standard prompting, we observe that: 1) As for the next-topic prediction (\textbf{hits@k}), thanks to the extensive knowledge across various topics, zero-shot LLMs can achieve competitive (Vicuna) or even better (ChatGPT) performance than the fine-tuned methods. 
2) As for the transition response generation, automatic evaluation metrics (\textbf{BLEU}, \textbf{METEOR}, \textbf{R-L})\footnote{
Note that the automatic evaluation of response generation is less reliable~\cite{acl21-otters}, as the same topic can be described in different ways rather than the reference response. We mainly discuss the topic shifting capability in terms of the performance on next topic prediction.} suggest that zero-shot models perform closely to fine-tuned methods in terms of lexical similarity with the reference response. 
3) One-shot ICL casts no positive impact on the performance and may even lead to worse results in next-topic prediction. This indicates that it is difficult for LLMs to enhance the topic shifting capability from limited demonstrations.

\paragraph{Only ProCoT prompting with one-shot demonstrations can improve the topic shifting capability.}
Without demonstrations, proactive and ProCoT prompts perform even worse than standard prompts, since LLMs may confuse about what kinds of topics are desired. For example, we observe a typical mistake that LLMs tend to analyse the next topics using questions, such as "\textit{What kind of food do you like?}", leading to a narrow topic for the next turn. 
With one-shot demonstrations, ChatGPT with proactive prompts continues to underperform compared to standard prompts when it comes to accurately predicting suitable topics towards the target. However, it is worth noting that only ProCoT prompts consistently show an improvement in the performance of all LLMs for next topic prediction.

\subsubsection{Dialogue-level Evaluation}

Table~\ref{tab:target_dial} shows the dialogue-level evaluation results on TGConv. We draw the following conclusions:

\paragraph{LLM-based dialogue systems tend to make aggressive topic transition.}   
The results demonstrate the effectiveness of LLMs in steering the conversation towards the designated target, with ChatGPT exhibiting nearly perfect success rates (\textbf{Succ.}). 
Compared with baselines, LLMs also excel in generating more coherent responses that align with the dialogue context (\textbf{Coh.}), showcasing their impressive abilities in context understanding and response generation. 
Furthermore, the analysis reveals that ChatGPT basically achieves the target topics within just three turns, suggesting its tendency to generate responses that aggressively involve the desired topic. Similar observations can be made with Vicuna using standard prompting.

\paragraph{ProCoT prompting enables a smoother topic transition of target-guided dialogues.} 
Under proactive prompting, the response coherency is improved by the topic planning. However, the success rate is negatively affected, which attributes to its drawback of next topic prediction discussed in Section~\ref{sec:turn}. 
Under ProCoT prompting, Vicuna effectively guide the conversation towards the designated target with a smoother (higher \textbf{Coh.}) and more engaging (higher \textbf{Turns}) conversation than using standard prompting. 
However, it still remains challenging for ChatGPT to perform a smooth topic transition. 
Case studies in Appendix~\ref{app:case-target} provide intuitive examples for illustrating these observations.

\begin{table}
\setlength{\belowcaptionskip}{5pt}
    \centering
    \setlength{\tabcolsep}{1mm}{
    \begin{adjustbox}{max width=0.48\textwidth}
    \begin{tabular}{lccccccccc}
    \toprule
       && & \multicolumn{3}{c}{\textbf{Easy Target}} & \multicolumn{3}{c}{\textbf{Hard Target}} \\
       \cmidrule(lr){4-6}\cmidrule(lr){7-9}
        Method & Shot& Prompt & Succ.(\%) & Turns & Coh.  & Succ.(\%) & Turns & Coh.  \\
    \midrule
     GPT2 &- &- &   22.3 &\underline{2.86}& 0.23& 17.3& \underline{2.94}& 0.21\\
     DKRN&- &- & 38.6 &4.24 &0.33 &21.7 &7.19 &0.31\\
     CKC &- &- &  41.9& 4.08& 0.35& 24.8& 6.88& 0.33 \\
     TopKG &- &- & 48.9& 3.95& 0.31& 27.3& 4.96& 0.33\\
     \textsc{Color} &- &- & \underline{66.3} & - & \underline{0.36} & \underline{30.1} & - & \underline{0.35}\\
     \midrule
     \multirow{6}{*}{Vicuna-13B} 
     & 0 & Standard & 63.0 & \textbf{2.63} & 0.43 & 62.5 & \textbf{2.45} & 0.39\\
        & 1 & Standard & 62.7 & 2.83 & 0.45 & \textbf{65.0} & 2.90 & 0.43\\
        & 0 & Proactive & 37.8 & 2.71&0.48& 35.6 & 2.56 & \textbf{0.55} \\
        & 1 & Proactive & 48.3 & 2.71 & 0.50 & 34.6 & 2.95 & 0.51 \\
        & 0 & ProCoT &65.2 &4.22&0.49& 54.9 & 4.17 & 0.45\\
        & 1 & ProCoT &\textbf{72.3} &3.55&\textbf{0.52}&59.8 & 3.81 & 0.48\\
     \midrule
     \multirow{6}{*}{ChatGPT} & 0 & Standard & \textbf{97.5}&\textbf{2.26} & 0.38 & \textbf{96.3} & 2.30 & 0.41\\
     & 1 & Standard & 96.3 & 2.42 & 0.42 & 93.5 & \textbf{2.28} & 0.38\\
        & 0 & Proactive & 85.9&3.20&\textbf{0.47}& 83.0 & 2.83 & \textbf{0.43}\\
        & 1 & Proactive & 90.7 & 2.86 & 0.36 & 86.2 & 2.94 & 0.31\\
        & 0 & ProCoT & 96.3 & 2.47 & 0.41 & 92.0 &2.29 & 0.34\\
        & 1 & ProCoT & 95.9 & 2.63 & 0.45 & 92.1 & 2.47 &0.39\\
     \bottomrule
    \end{tabular}
    \end{adjustbox}}
    \caption{Dialogue-level evaluation results on target-guided dialogues. }
    \label{tab:target_dial}
    \vspace{-0.3cm}
\end{table}

\begin{table}[]
\setlength{\belowcaptionskip}{0pt}
    \centering
    \setlength{\tabcolsep}{1mm}{
    \begin{adjustbox}{max width=0.35\textwidth}
    \begin{tabular}{lccccc}
    \toprule
    && \multicolumn{2}{c}{Easy Target} & \multicolumn{2}{c}{Hard Target} \\
    \cmidrule(lr){3-4}\cmidrule(lr){5-6}
    Method    & Prompt  & G-Coh. & Effect. & G-Coh. & Effect. \\
    \midrule
    TopKG & - & 1.42& 1.24&1.21&1.10\\
    \midrule
    \multirow{3}{*}{Vicuna-13B}    & Standard &1.37 & 1.60 & 1.20 & 1.49 \\
       & Proactive & 1.51 & 1.27&1.26&1.23\\
      & ProCoT & \textbf{1.57}& \textbf{1.70} & \textbf{1.35}& \textbf{1.59}\\
    \midrule
    \multirow{3}{*}{ChatGPT}   & Standard & 0.97 & \textbf{1.92} & 0.84 & \textbf{1.89}\\
       & Proactive & \textbf{1.24} & 1.77 & 1.12 & 1.68\\
       & ProCoT & 1.20 & 1.90 & \textbf{1.14} & 1.85\\
       \bottomrule
    \end{tabular}
    \end{adjustbox}}
    \caption{Human evaluation on target-guided dialogues. All reported methods are under the one-shot setting.}
    \label{tab:human-target}
    \vspace{-0.5cm}
\end{table}

\subsubsection{Human Evaluation}
Table~\ref{tab:human-target} presents the human evaluation results on TGConv. 
Compared with TopKG, LLMs demonstrate remarkable efficiency in achieving the designated target (\textbf{Effect.}). However, the global coherence (\textbf{G-Coh.}) of the generated dialogues by ChatGPT is quite low, which may harm the user engagement and experience during the conversation. 
Thus, the proficiency of controllable generation in LLMs is a double-edged sword for target-guided dialogues. 
\textbf{The key challenge of LLMs is how to guarantee the topical smoothness and coherence of the generated transition responses. }

\begin{table}
\setlength{\belowcaptionskip}{0pt}
    \centering
    \setlength{\tabcolsep}{1mm}{
    \begin{adjustbox}{max width=0.48\textwidth}
    \begin{tabular}{lcccccccc}
    \toprule
       && & \multicolumn{2}{c}{\textbf{Nego. Strategy}} & \multicolumn{2}{c}{\textbf{Dial. Act}} &\multicolumn{2}{c}{\textbf{Resp. Gen.}} \\
       \cmidrule(lr){4-5}\cmidrule(lr){6-7}\cmidrule(lr){8-9}
        Method & Shot& Prompt &{F1}& {AUC} & {F1}& {AUC} & BLEU & {BERTScore}\\
    \midrule
     FeHED    & - & - &  17.6  &55.8  &20.6  &76.9 & 23.7 &27.0\\
     HED+RNN    & - & - &  23.2 & 65.3  &33.0  &83.1 &22.5&22.8\\
     HED+TFM & - & - & \underline{26.3} &\underline{68.2} &32.5 &85.6 &24.4&27.7\\
     \textsc{DialoGraph} & - & - & 26.1  &68.1 &\underline{33.4} &\underline{85.6} &\underline{24.7}& \underline{28.1}\\
     \midrule
     \multirow{6}{*}{Vicuna-13B} & 0 & Standard &-&-&-&-&1.7&-14.0\\
     & 1 & Standard &-&-&-&-&1.9&-2.8\\
        & 0 & Proactive & \textbf{20.6}&\textbf{51.1}&4.2&50.3&2.3&-7.0\\
        & 1 & Proactive &15.2&50.0&6.7&50.8&2.6&-0.9\\
        & 0 & ProCoT & 19.0&49.7&3.6&50.3&2.6&-6.2\\
        & 1 & ProCoT & 17.8&48.9&\textbf{7.7}&\textbf{52.5}&\textbf{2.6}&\textbf{-0.9}\\
     \midrule
     \multirow{6}{*}{ChatGPT} & 0 & Standard &-&-&-&-&2.3&-4.3\\
     & 1 & Standard &-&-&-&-&3.1&0.7\\
        & 0 & Proactive & 12.8&51.3&13.3&56.3&\textbf{4.2}&1.3\\
        & 1 & Proactive & 13.7&50.9&12.0&54.9&3.9&\textbf{2.9}\\
        & 0 & ProCoT &10.8&50.4&10.1&54.2&3.7&-0.9\\
        & 1 & ProCoT &\textbf{15.1}&\textbf{55.5}&\textbf{16.3}&\textbf{58.2}&3.9&1.6\\
     \bottomrule
    \end{tabular}
    \end{adjustbox}}
    \caption{Evaluation results on Negotiation Strategy Prediction, Dialogue Act Prediction, and Response Generation. }
    \label{tab:nego_turn}
\vspace{-0.4cm}
\end{table}

\subsection{Non-collaborative Dialogues}\label{sec:non-colla}
Unlike collaborative task-oriented dialogue settings~\cite{tod-survey}, where the user and the system work together to reach a common goal (\textit{e.g.}, booking hotels), in non-collaborative dialogues, the user and the system have a conflict of interest but aim to strategically communicate to reach an agreement (\textit{e.g.}, negotiation)~\cite{negotiate-survey}. The system is required to leverage a series of proactive strategies to reach an agreement favorable to itself, instead of passively following the user's intents.

\subsubsection{Problem Definition}
Given the dialogue history $\mathcal{C}=\{u_1,...,u_{t-1}\}$ and the dialogue background $\mathcal{D}$, the goal is to generate a response $u_t$ with appropriate dialogue strategy $a_t$ that can lead to a consensus between the system and user. 
A set of dialogue strategies $\mathcal{A}$ is pre-defined for prediction. 
Based on different applications, the dialogue strategy can be coarse dialogue act labels or fine-grained strategy labels. The dialogue background includes the system's goal and the related grounded information, such as item descriptions in bargain negotiation~\cite{emnlp18-negotiate} and user profile in persuasion dialogues~\cite{acl19-persuasion}.

\subsubsection{Experimental Setups}
\paragraph{Datasets.} 
We use the \textbf{CraigslistBargain} dataset \cite{emnlp18-negotiate} for evaluating the capability of strategically handling non-collaboration in LLM-based dialogue systems. The dataset was created under the bargain negotiation setting where the buyer and the seller are negotiating the price of an item on sale. 
Details can be found in Appendix~\ref{app:dataset}. 

\paragraph{Automatic Evaluation Metrics.}
Following the previous study~\cite{iclr21-negotiate-strategy}, we conduct a comprehensive evaluation over three subtasks, including negotiation strategy prediction, dialogue act prediction, and response generation. 
We report the F1 and ROC AUC scores for strategy prediction and dialogue act prediction, where the former one is a multi-label prediction problem. 
For the response generation, we adopt BLEU score and BERTScore~\cite{bertscore} for evaluation. 

\begin{table}[]
\setlength{\belowcaptionskip}{0pt}
    \centering
    \setlength{\tabcolsep}{1mm}{
    \begin{adjustbox}{max width=0.38\textwidth}
    \begin{tabular}{lcccccc}
    \toprule
    Metric    & Standard   & Proactive & ProCoT & Gold\\
    \midrule
    Persuasive & 1.24 & 1.28 & 1.43 & \textbf{1.54}\\
    Coherent & 1.56 & 1.66 & \textbf{1.74} & 1.69\\
    Natural & 1.94 & 1.82 & 1.89 & \textbf{1.97}\\
    \midrule
    Win Rates\\
    ~ - vs. Standard & - & 0.22 & 0.24 & \textbf{0.42}\\
    ~ - vs. Proactive & 0.25 & - & 0.31 & \textbf{0.45}\\
    ~ - vs. ProCoT & 0.20 & 0.18 & - & \textbf{0.34}\\
    ~ - vs. Gold & 0.19 & 0.09 & 0.23 & -\\
    \midrule
    Sale-to-List Ratio & 0.48 & 0.43 & 0.54 & \textbf{0.64}\\
       \bottomrule
    \end{tabular}
    \end{adjustbox}}
    \caption{Human evaluation on non-collaborative dialogues. All reported methods are based on ChatGPT under the one-shot setting. \textbf{Gold} denotes that we instruct the LLMs to generate responses conditioned on the reference dialogue acts and negotiation strategies.}
    \label{tab:human-noncollab}
    \vspace{-0.3cm}
\end{table}

\paragraph{Human Evaluation Metrics.}
Following~\citet{iclr21-negotiate-strategy}, we also conduct human evaluation on 100 randomly sampled dialogues with both subjective and objective human judgement. 
As for the subjective judgement, annotators are asked to score [0,1,2] on how persuasive, coherent, and natural the generated response is. 

We further pair the generated responses from each prompting scheme, including Standard, Proactive, ProCoT, and Ground-truth (GT), with the corresponding responses from each of the other prompting scheme to compute the overall win rates between each pair. 

As for the objective human judgement, we adopt the sale-to-list ratio (SL\%)~\cite{iclr21-negotiate-strategy,eacl21-persuasion-strategy} as an indicator for explicitly measuring the negotiation inclination in the generated response: 
\begin{equation}
    SL\% = \frac{\text{bargain price} - \text{buyer target price}}{\text{listed price} - \text{buyer target price}},
\end{equation}
where the bargain price is the price that the seller would like to sell the item at the current turn. The lower the SL\%, the more compromise the seller have made.

To sum up, the instructions provided for annotators are as follows: 
\begin{itemize}[leftmargin=*]
    \item Persuasive: Whether the seller is persuasive in bargaining the price. 
    \item Coherent: Whether the seller's responses are on topic and in line with the conversation history. 
    \item Natural: Whether the seller is human-like. 
    \item Bargain Price: What is the current bargain price from the seller's side. 
    \item Win: Assume you are the seller. Which dialogue system you would like to use for bargain the price with the buyer (Win/Tie/Lose).  
\end{itemize}

\begin{figure*}
\setlength{\abovecaptionskip}{0pt}   
\setlength{\belowcaptionskip}{0pt}
\centering
\includegraphics[width=0.93\textwidth]{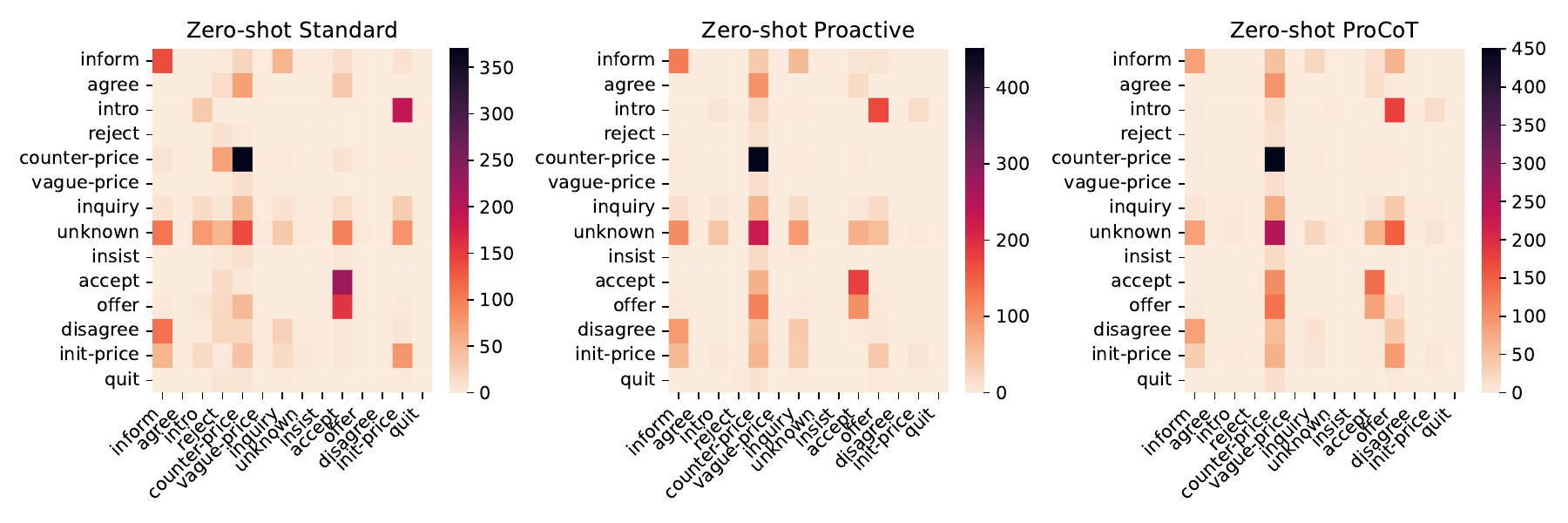}
\caption{Heatmaps on the relationships between target and predicted dialogue acts. As no dialogue act is predicted in standard prompting, a dialogue act classifier is trained to identify the dialogue act of the generated response.}
\label{dialact}
\vspace{-0.5cm}
\end{figure*}

\paragraph{Usage of LLMs \& Prompting Schemes.}
The adopted LLMs are the same, but the maximum number of new tokens is set to be 256, as there are more information needed to be generated, including negotiation strategies and dialogue acts.

\paragraph{Baselines.}
We compare several fine-tuned SOTA baselines for negotiation dialogues, including FeHED~\cite{iclr20-non-collab}, HED+RNN/TFM, and \textsc{DialoGraph} \cite{iclr21-negotiate-strategy}. 

\subsubsection{Experimental Results}
Table~\ref{tab:nego_turn} and Table~\ref{tab:human-noncollab} present the results with automatic and human evaluation metrics, respectively. There are several notable findings as follows: 

\paragraph{LLM-based dialogue systems fail to predict appropriate negotiation strategies and dialogue acts.}
Table~\ref{tab:nego_turn} shows that failures on strategy learning further result in a poor performance of response generation. Specifically, ChatGPT generally performs better than Vicuna in strategy learning. 
Although both proactive and ProCoT prompting schemes can slightly improve the final performance of response generation, there is still a large gap from fine-tuned methods according to automatic evaluation metrics.

\paragraph{The key challenge of LLMs in handling non-collaborative dialogues is how to effectively optimize the strategy planning.} 
Table~\ref{tab:human-noncollab} shows that the generated responses conditioned on reference strategies are more favorable (\textbf{Win Rates}). 
In specific, ChatGPT guarantees a high score on the human-like response generation (\textbf{Natural}). 
With the ProCoT, the generated responses are more coherent to the conversation history (\textbf{Coherent}), which can also be observed from the case study in Appendix~\ref{app:case-non-collab}. 
However, compared with prompting with reference strategies, all the other prompting schemes fall short of generating persuasive responses for negotiation (\textbf{Persuasive}), indicating their shortcomings on strategy learning. 
This is also validated by the objective judgement on \textbf{Sale-to-List Ratio}, which shows that ChatGPT can reach a better deal for itself when being conditioned on reference strategies. 
Similarly, \citet{acl23-prompt-mixed} empirically show that, given the optimal planned strategy, ChatGPT achieves strong performance on controllable response generation in some other strategy-based dialogues.

\subsubsection{Analysis of Strategy Learning}
Figure~\ref{dialact} presents the analysis of the relationships between the target and predicted dialogue acts by ChatGPT. 
As for the standard prompting, we observe two typical mistakes: 1) The system tends to propose the initial bargain price (\texttt{init-price}), instead of greetings (\texttt{intro}) and waiting for the buyer to initialize the bargain. 2) The system often directly accepts the buyer's offer (\texttt{accept}) when it is supposed to offer another price for negotiation (\texttt{offer}). 
This also explains why the \textbf{Sale-to-List Ratio} is relatively low when using standard prompting in Table~\ref{tab:human-noncollab}. 
On the other hand, Proactive and ProCoT prompting share similar patterns of mistakes, where ChatGPT tends to propose a counter price (\texttt{counter-price}) to negotiate with the buyer.

Appendix~\ref{app:strategy} presents the analysis of the distribution of selected strategies by ChatGPT. In the reference responses, the seller often shows positive/negative sentiment to negotiate with the buyer. However, ChatGPT inclines to adopt conservative or concessionary strategies, such as using hedge words, show gratitude, or propose a counter price. 

Overall, we conclude that \textbf{ChatGPT tends to make compromise with the buyer during the negotiation, rather than strategically taking actions to maximize its own benefit}. 

\section{Conclusion}
In this work, we conduct the first comprehensive evaluation on the capability of LLM-based dialogue systems in handling proactive dialogues, including clarification, target-guided, and non-collaborative dialogues. To enhance the proactivity of LLM-based dialogue systems, we propose a proactive chain-of-thought prompting scheme that triggers the reasoning and planning capability of LLMs. 
The empirical analysis sheds light on the potentials of LLMs for proactive dialogues: 1) ProCoT largely enhances the originally poor performance of LLMs in asking clarification questions, but still limits in handling domain-specific applications. 2) LLM-based dialogue systems perform aggressive topic shifting towards the designated target, while ProCoT enables the topic planning to be smoother. 3) Despite the strength on controllable response generation, the capability of strategy learning and planning is a key challenge for LLMs in handling non-collaborative dialogues. 

\section*{Acknowledgement}
This research is supported by NExT Research Center.

\section*{Limitation}
In this section, we discuss the limitations of this work from the following perspectives:

\paragraph{Sensitivity of Prompts}
Similar to other studies on prompting LLMs for dialogue applications~\cite{acl23-prompt-ood,acl23-prompt-mixed,acl23-robust}, the evaluation results are likely to be sensitive to the choice of prompts. Besides, it is also likely that the designed prompts are not the optimal ones for the concerned problem. 
In fact, prompt sensitivity and optimality themselves are valuable research problems in dialogue systems, which can be further investigated in the future studies. 
To facilitate the reproducibility of this work, we will release all the prompts used in the experiments and provide detailed descriptions about the designs of each prompting scheme in Appendix~\ref{app:example}. The code and data will be released via \url{https://github.com/dengyang17/LLM-Proactive}. 

\paragraph{Financial and Computational Cost of LLMs}
It is financially expensive to call the API of commercial LLMs for experiments. In our experiments, it costs about \$120 to call the OpenAI API for getting all the experimental results of ChatGPT. 
On the other hand, it is computationally expensive to conduct experiments with open-source LLMs in local machines. 
In our experiments, we choose Vicuna 13B as the open-source LLM for evaluation, which can be adapted to NVIDIA DGX-1 V100 32G for inference. 
If more budgets and better experimental environment are permitted, it would be great to evaluate how other larger LLMs performs in the concerned proactive dialogue problems, such as GPT-4, LLaMA/Vicuna 65B, etc. 

\paragraph{Capability of Planning and Decision Making} 
The proposed ProCoT prompting scheme can be regarded as a preliminary attempt at triggering the capability of planning and decision making from LLM-based dialogue systems. Compared with fine-tuned methods, such ability of LLMs is still weak as we learn from the empirical analysis. 
Moreover, simply prompting LLMs to be proactive may fall short of handling decision making under dynamic environments in real-world applications.  
It is worth studying how LLM-based dialogue systems handle the proactive dialogue problems in an interactive setting with more diverse user simulation~\cite{sigir22-proactive,negotiate-selfplay}.

\bibliography{custom}
\bibliographystyle{acl_natbib}

\appendix

\section{Details of Datasets}
\label{app:dataset}
In the experiment, we adopt the test sets from five datasets for evaluation, including Abg-CoQA~\cite{abg-coqa}, PACIFIC~\cite{pacific}, Otters~\cite{acl21-otters}, TGConv~\cite{coling22-topkg}, and CraigslistBargain~\cite{emnlp18-negotiate}. 
Detailed descriptions of each dataset are as follows:
\begin{itemize}[leftmargin=*]
    \item Abg-CoQA\footnote{\url{https://github.com/MeiqiGuo/AKBC2021-Abg-CoQA/tree/main/abg-coqa}} is constructed based on the CoQA dataset~\cite{coqa} by truncating a partial conversation from the full conversation and selecting  ambiguous questions. 
    \item PACIFIC\footnote{\url{https://github.com/dengyang17/PACIFIC/tree/main/data/pacific}. Since the labels in the test set is not publicly released, we adopt the validation set for evaluation.} is constructed based on the TAT-QA dataset~\cite{tat-qa}, an question answering dataset in the financial domain, whose contexts contain a hybrid of tables and texts. \citet{pacific} rewrite the questions to be ambiguous for introducing clarification turns in the conversation. 
    \item OTTers is a next-turn target-oriented dialogue dataset, which requires the agent proactively generate a transition utterance to approach the designated target. We adopt the processed version\footnote{\url{https://github.com/yyyyyyzt/topkgchat}} by \citet{coling22-topkg} for evaluation. The topic is represented as a set of topical keywords. 
    \item TGConv is constructed based on ConvAI2~\cite{convai2} and is split to two settings, including "easy-to-reach" and "hard-to-reach". The topic is also represented as a set of topical keywords. 
    \item CraigslistBargain was created in a negotiation setting where two crowdsourced workers play the roles of the buyer and the seller to bargain the price of an item. We adopt the processed version\footnote{\url{https://github.com/rishabhjoshi/DialoGraph\_ICLR21/tree/main}} by \citet{iclr21-negotiate-strategy} for evaluation, which assigns 10 dialogue acts and 21 negotiation strategies to the utterances. 
\end{itemize}

\begin{table}[]
    \centering
    \begin{adjustbox}{max width=0.4\textwidth}
    \begin{tabular}{lrrr}
    \toprule
    &&Avg. & Avg.\\
      Dataset   & \#Dialog & \#Turns &\#Words \\
      \midrule
       Abg-CoQA  & 1055 & 5.04 & 4.87\\
       PACIFIC & 278 & 7.10 & 6.80\\
       OTTers & 372 & 3.00 & 9.19 \\
       TGConv & 751 & 8.97 & 12.47\\
       CraigslistBargain & 567 & 9.78 & 11.65\\
       \bottomrule
    \end{tabular}
    \end{adjustbox}
    \caption{Statistics of datasets.}
    \label{tab:dataset}
\end{table}

\begin{table}
\fontsize{8.5}{10}\selectfont
    \centering
    \begin{adjustbox}{max width=0.5\textwidth}
    \begin{tabular}{p{1.5cm}|p{5.2cm}}
    \toprule
       Under-specified & Q: What is the 2019 average defined schemes?\\
       Clarification & G: What kind of defined schemes are you\\
        & asking about?\\
        &    P: Could you please clarify what you mean by "average defined schemes"?\\
       \midrule
       Wrong ~~~~ Aspect & Q: What was the increase / (decrease) in the cost? \\
        & G: Which period are you asking about?  \\
        & P: Which cost are you referring to? \\
       \midrule
       Over-specified & Q: How much amount of goodwill acquisitions was done in 2019? \\
       Clarification  & G: What kind of goodwill acquisitions are you asking about?\\
        &   P: Are you asking about the total amount of goodwill activity or specifically about acquisitions in 2019?\\
       \midrule
       Generation &Q: How much was the average operating \\
       Error & income?\\
        & G: Which period are you asking about? \\
       &P: The document does not provide information on the average operating income. Therefore, the question cannot be answered.\\
       \bottomrule
    \end{tabular}
    \end{adjustbox}
    \caption{Error Analysis (Q: User Query, G: Ground-truth, P: Prediction).}
    \label{tab:error}
\end{table}

\begin{figure*}
\centering
\includegraphics[width=\textwidth]{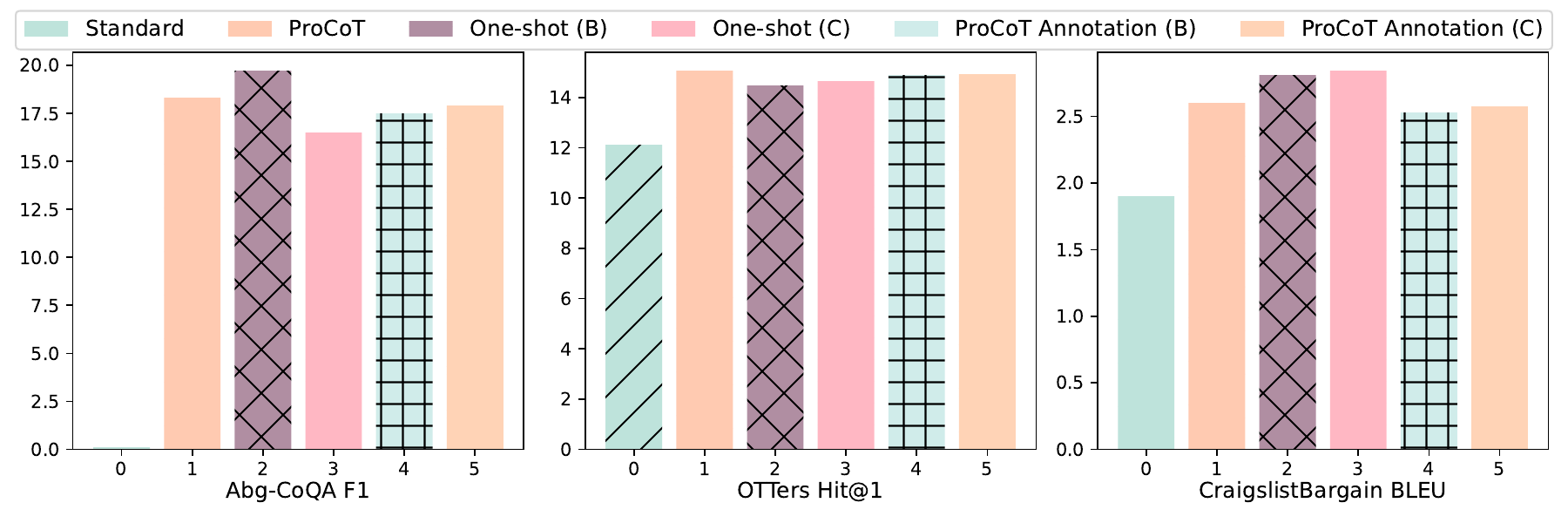}
\caption{Performance in terms of different prompts, including different one-shot examples and different ProCoT annotations. The reported results are based on Vicuna-13B. }
\label{prompt_analysis}
\end{figure*}

\section{Error Analysis Details for Clarification Dialogues}\label{app:error}
As shown in Table~\ref{tab:error}, we categorize these failure cases into the following four groups:  
\begin{itemize}[leftmargin=*]
    \item \textit{Wrong Aspect}: The model generates a question for clarifying a wrong aspect of the user query. 
    \item \textit{Under-specified Clarification}: The model generates an under-specified clarification question, where the requested information is too general so that it would be difficult for the user to provide feedbacks.  
    \item \textit{Over-specified Clarification}: The model generates an over-specified clarification question, where the requested information is already clear in the user query. 
    \item \textit{Generation Error}: Although the model identifies the need for clarification, but it doesn't generate the output as the required format, such as no clarification question. 
\end{itemize}

\begin{figure}[t]
\centering
\includegraphics[width=0.48\textwidth]{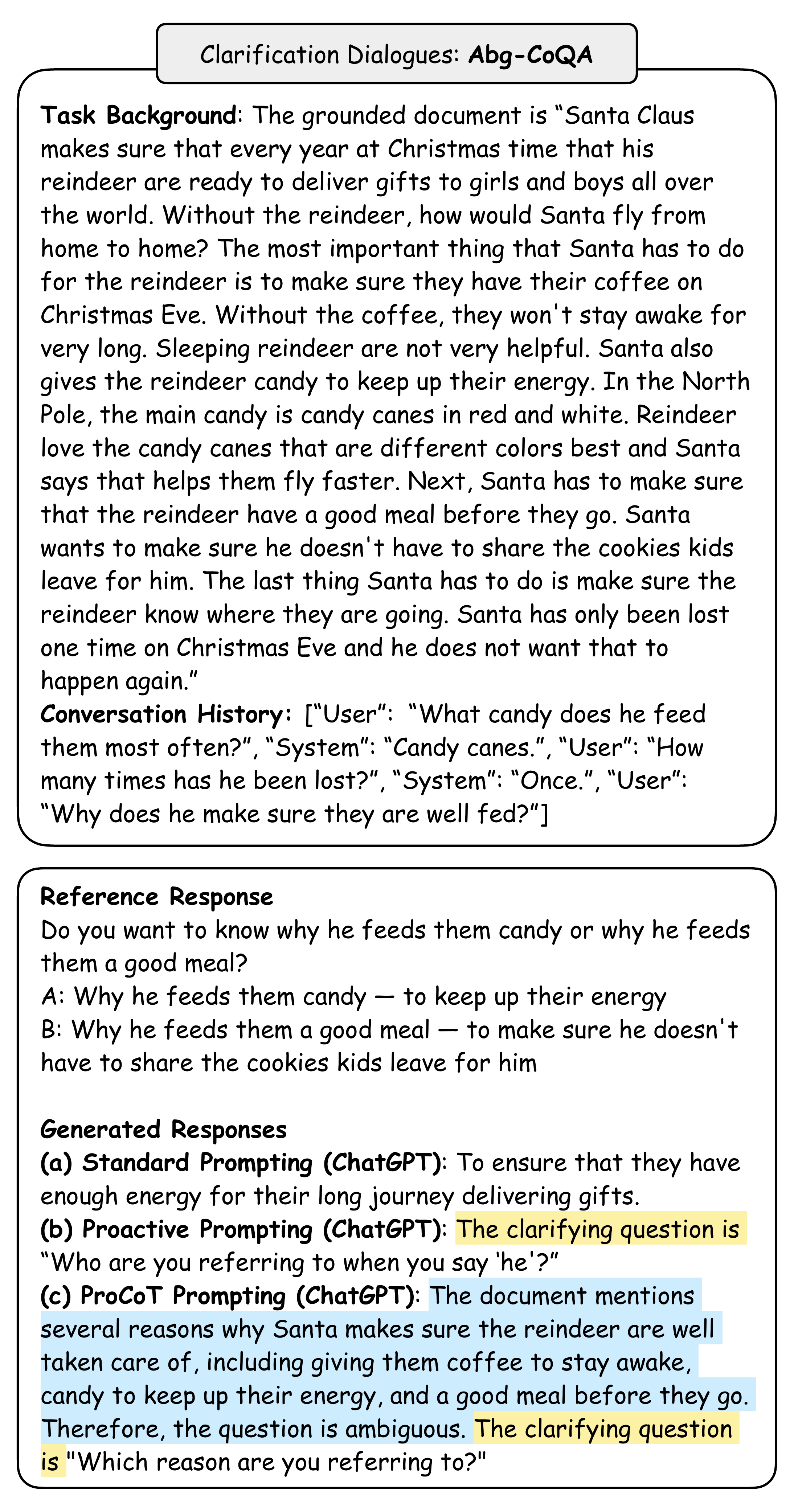}
\caption{Case study from the Abg-CoQA dataset.}
\label{abgcoqa-case}
\end{figure}

\begin{figure}[t]
\centering
\includegraphics[width=0.48\textwidth]{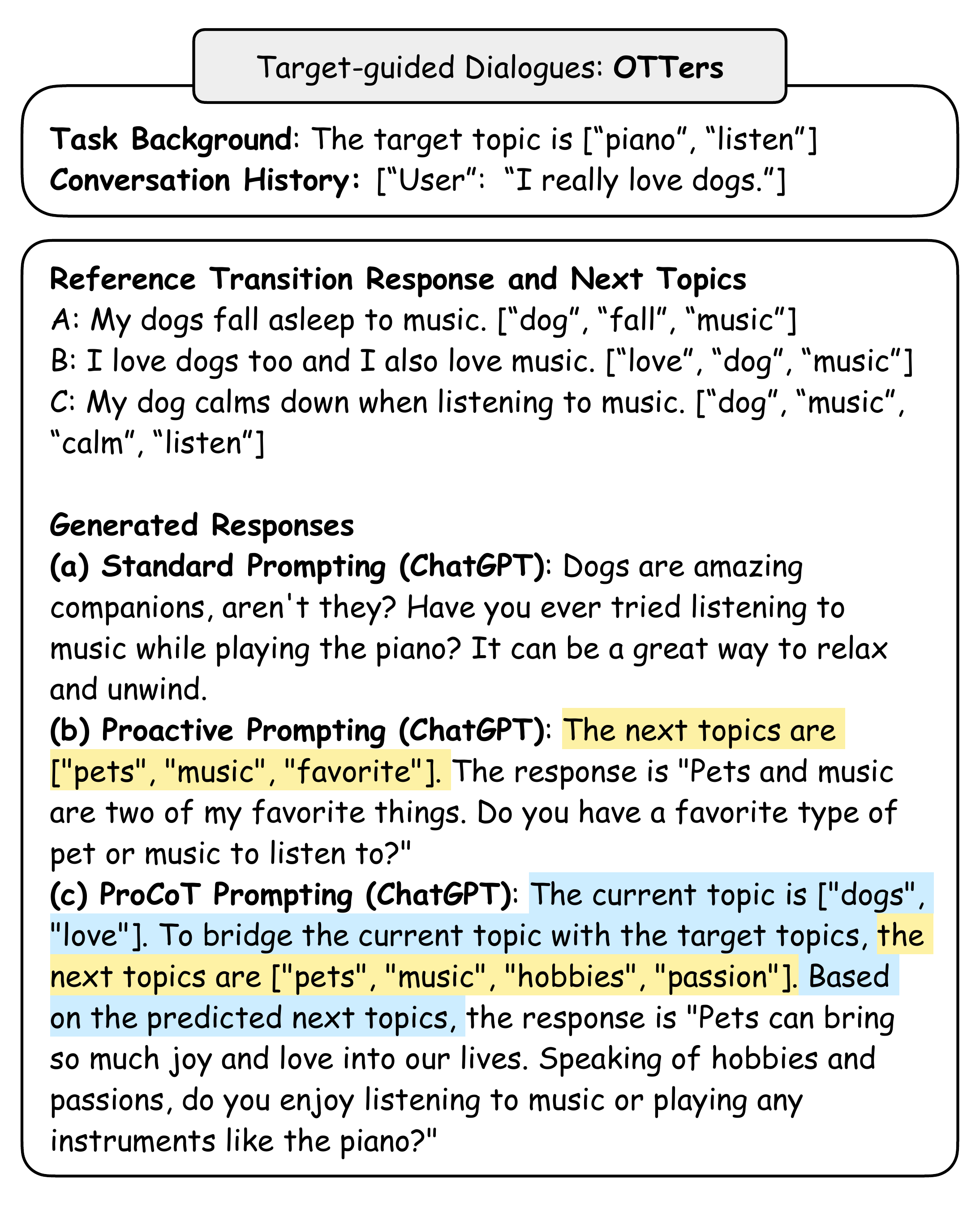}
\caption{Case study from the OTTers dataset.}
\label{otters-case}
\end{figure}

\begin{figure}[t]
\centering
\includegraphics[width=0.48\textwidth]{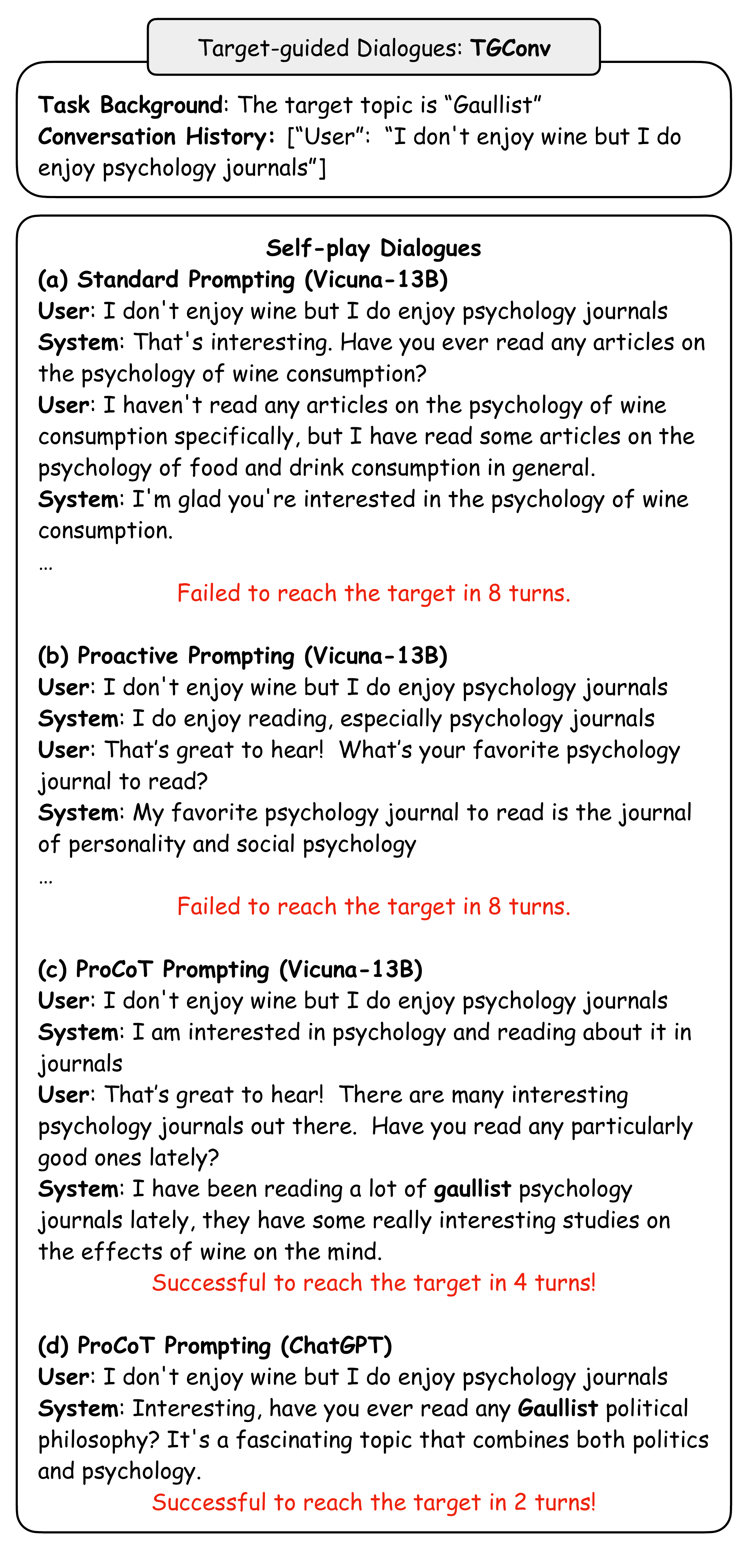}
\caption{Case study from the TGConv dataset.}
\label{topkg-case}
\end{figure}

\begin{figure}[!t]
\centering
\includegraphics[width=0.48\textwidth]{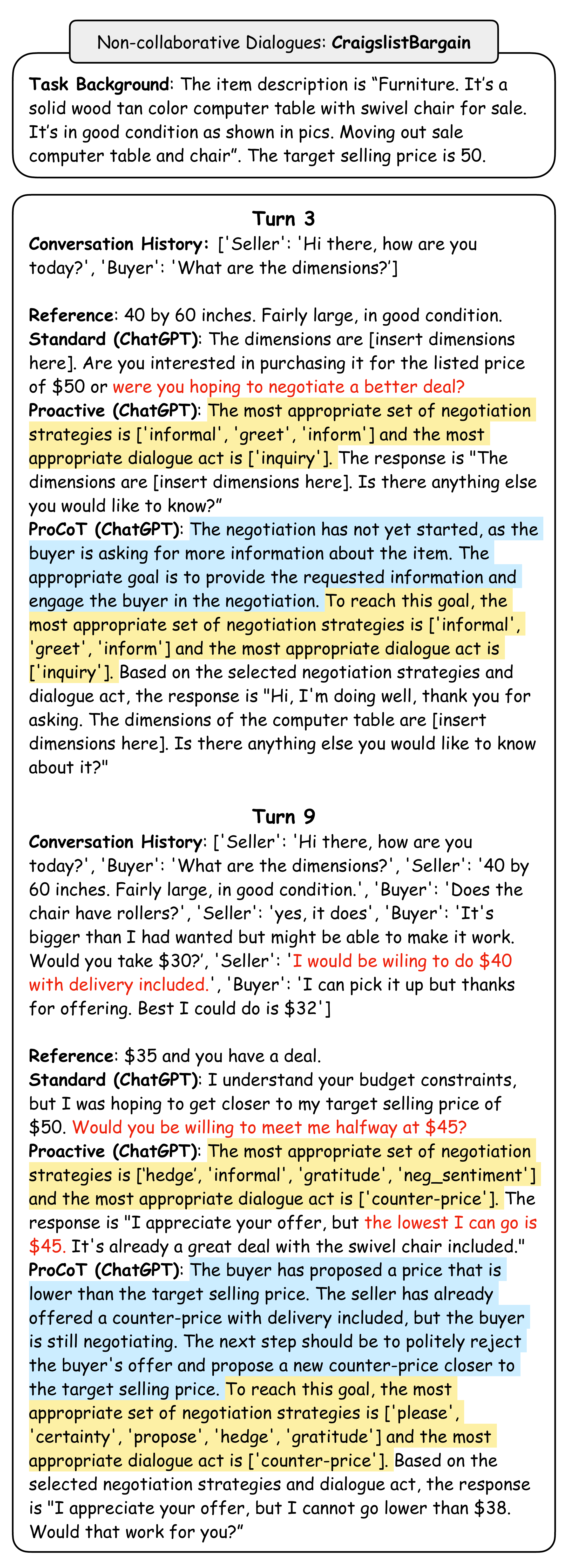}
\caption{Case study from the CraigslistBargain dataset.}
\label{cb-case}
\end{figure}

\section{Designs of Prompts}\label{app:example}
Table~\ref{tab:cq_example}, \ref{tab:tg_example}, and \ref{tab:nc_example} present the example of prompts for clarification, target-guided, non-collaborative dialogues, respectively. 
As for the zero-shot setting, the overall prompt is composed by the task instruction and the sample. 
As for the few-shot setting, the overall prompt is composed by the task instruction, a number of samples with demonstrations, and the test sample. 

In particular, we clarify several questions regarding the prompt designs as follows:

\paragraph{How to construct the task instructions?} The task instructions first follow the problem definition for each proactive dialogue problem. Then, similar to other studies on applying LLMs for different tasks~\cite{chatgpt-ie,mmmeval-chatgpt}, we further instruct the LLMs to generate the response following the desired output format for evaluation. 

\paragraph{How to choose the one-shot sample?} Due to the input length limitation of LLMs, we could only adopt one-shot in-context learning (ICL). 
In order to testify the sensitivity of the choice of the one-shot sample, we report the results with three different one-shot samples in Figure~\ref{prompt_analysis}. 
Despite the variance among different one-shot examples as expected when using exemplar-based ICL~\cite{naacl21-prompt,icml21-fewshot}, the observations of adopting ProCoT for different proactive dialogues remain unchanged. 
Since the variance of one-shot ICL is inevitable, we simply adopt the first dialogue sample in the original training set of each dataset as the one-shot sample for facilitating reproducibility. 

\paragraph{How to construct the demonstration of proactive chain-of-thoughts?} 
The demonstration of proactive chain-of-thoughts is written by human annotators, which represents their own chain-of-thoughts of the planning or decision making for the proactive dialogues. 
Following \citet{cot}, we also report the results with the demonstrations from three different annotations. 
As shown in Figure~\ref{prompt_analysis}, it can be observed that the variance is much smaller than the one-shot sample. We adopt the best ProCoT annotation for each dataset in the evaluation.

\section{Case Study}
In order to intuitively compare the three prompting schemes, we conduct case studies on the generated responses regarding three proactive dialogue problems. 
\subsection{Clarification Dialogues}\label{app:case-clari}

\begin{figure*}
\centering
\includegraphics[width=\textwidth]{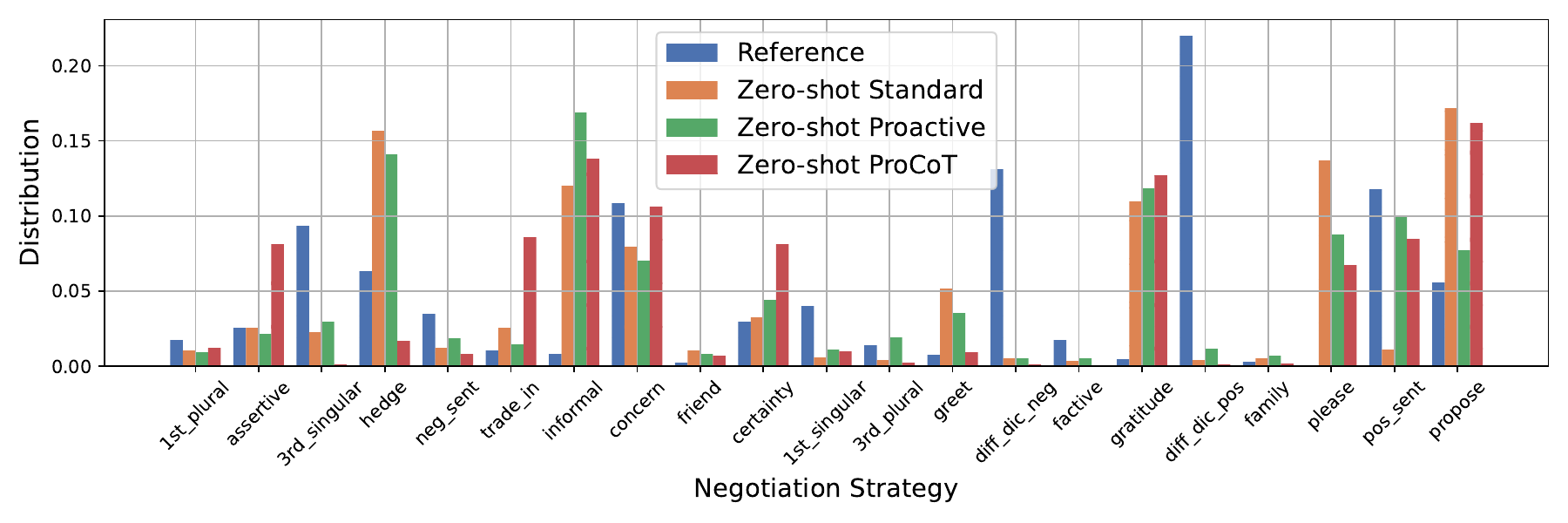}
\caption{Distribution of selected negotiation strategies. Similarly, a negotiation strategy classifier is trained to identify the negotiation strategies of the generated response in standard prompting.}
\label{strategy}
\end{figure*}

Figure~\ref{abgcoqa-case} presents a case study from the Abg-CoQA dataset. There are two possible answers for the ambiguous user question in this case. When using standard prompting (a), ChatGPT generates a response based on a random guess, potentially resulting in an answer that may not align with the user's desired outcome. 
When using proactive prompting (b), although ChatGPT decides to ask a clarification question, the concerned aspect of clarification is not the focus of this conversation. 
ProCoT prompting (c) first provides a thorough analysis to identify the underlying reasons for the ambiguity in the user's question. This analysis serves as the foundation for generating an effective clarifying question, which helps to clarify and disambiguate the user's intended meaning.

\subsection{Target-guided Dialogues}\label{app:case-target}

Figure~\ref{otters-case} presents a case study from the OTTers dataset, where the target topical keywords include "piano" and "listen", and the system is required to generate a transition response to approach the target topics. 
It can be observed that the current topics in the user utterance, \textit{i.e.,} "dog", is completely different from the target topics. 
When using standard prompting, ChatGPT leverages its overwhelming ability of controllable generation to generate the response with aggressive topic transition. 
Despite its fluency, it is not a natural utterance with logical or topical coherency. 
When using proactive prompting, although the predicted next topics are less aggressive than the standard prompting, the generated transition response just blends the current topics and the next topics together without naturally connecting the topics. 
Finally, ChatGPT with the ProCoT prompting generates a relatively smoother transition response to bridge the current topic with the target topic through planning about the topic shifting process.

Figure~\ref{topkg-case} presents a case study from the TGConv dataset, where the hard-to-reach target of this case is "Gaullist", and the system is required to reach this target within 8 turns of conversations under the self-play simulation. 
As for Vicuna, it is struggled to lead the conversation towards this hard-to-reach target, when using standard and proactive prompting. ProCoT prompting enables Vicuna to effectively and smoothly drive the conversation towards the designated target. 
In addition, owing to the powerful capability of controllable text generation, ChatGPT directly responds with the target topic to the initial user utterance. However, the topic transition is relatively aggressive, which might downgrade the user engagement or experience during the conversation.

\subsection{Non-collaborative Dialogues}\label{app:case-non-collab}

Figure~\ref{cb-case} presents a case study from the CraigslistBargain dataset, where the system plays the seller role to bargain with the buyer. At turn 3, even though the buyer just inquires about the item information without showing the bargain intention, ChatGPT with standard prompting tends to initiate the negotiation, which may put the seller in a disadvantageous position. Proactive and ProCoT prompting enable the dialogue act and strategy prediction of the next response. Especially for the analysis of the current negotiation status, ProCoT points out that the negotiation has not yet started. 

At turn 9, we observe that the seller has already lowered down the bargain price to \$40 in a previous turn. Without the reasoning and planning process, ChatGPT with standard and proactive prompting generates the response with contradictory statement, \textit{i.e.,} propose a higher counter price (\$45) for bargain, which is unreasonable in negotiation dialogues. With proactive CoTs, ChatGPT effectively summarizes the current negotiation progress and makes a better decision on the next negotiation goal.

\section{Analysis of Strategy Learning (Cont.)}\label{app:strategy}

Figure~\ref{strategy} presents the analysis of the distribution of selected strategies by ChatGPT. 
As for the reference responses, we observe that the seller tends to express their positive/negative sentiment as well as negotiate in a positive/negative manner. 
Differently, ChatGPT with standard and proactive prompting prefers to use hedge words or polite expressions (\textit{e.g.,} please and gratitude), indicating that ChatGPT essentially plays a nice role in negotiation. 
ChatGPT with ProCoT prompting makes more decisions to use assertive words or trade in, compared with other distributions. This shows that ProCoT can enable ChatGPT to involve certain negotiation strategies.

\begin{table*}[!t]
    \centering
    \begin{tabular}{p{0.97\textwidth}}
    \toprule
    \textbf{\textit{Clarification Dialogues}}     \\
    \midrule
    \textbf{Standard Prompting}: Given the document and the conversation history, generate the response.      \\
    \midrule
    \textbf{Proactive Prompting}: Given the document and the conversation history, answer the question or ask a clarifying question. The response should start with "The answer is" or "The clarifying question is".  \\
    \midrule
    \textbf{ProCoT Prompting}: Given the document and the conversation history, first identify whether the question is ambiguous or not. If it is ambiguous, ask a clarifying question. If it is not ambiguous, answer the question. The response should start with the ambiguity analysis of the question and then follow by "Therefore, the question is not ambiguous. The answer is" or "Therefore, the question is ambiguous. The clarifying question is". \\
    \midrule
    \makecell[l]{\textbf{Sample}: \\
    \multicolumn{1}{p{0.95\textwidth}}{Document: "Angie went to the library with her mother. First she had to turn in the books she was returning at the return desk. They said hello to the man there. He took their books. Then they went into the adult reading room. Angie sat in a brown chair at the table. She made a drawing of her mother. Her mother found a large red book. Then they went to the Mystery section. Angie sat in a blue chair. She drew a picture of her brother. Her mother found the book. It was a green book. Finally it was time to go to the children's room. It was Story Hour. Miss Hudson was there to read to all the children. She read a book about friendship. After the story Angie sat in the red chair and began drawing. They were drawing pictures of friends. Angie drew a picture of her best friend Lilly. Miss Hudson hung the pictures on the wall. Then Angie and her mother picked out 8 books to read at home. They checked the books out and went home."}\\
    \multicolumn{1}{p{0.95\textwidth}}{Conversation history: [“User”: “What did she draw?”, “System”: “Her mother”, “User”: “What did her mother find?”, “System”: “The book”]}\\
    \multicolumn{1}{p{0.95\textwidth}}{Question: “What color was it?”}}\\
    \midrule
    \textbf{Demonstration (Standard)}: Do you mean the first book?\\
    \midrule
    \textbf{Demonstration (Proactive)}: The clarifying question is "Do you mean the first book?"\\
    \midrule
    \textbf{Demonstration (ProCoT)}: There are two books that book that Angie's mother found. It is uncertain which book is referred to. Therefore, the question is ambiguous. The clarifying question is "Do you mean the first book?"\\
    \bottomrule
    \end{tabular}
    \caption{Examples of prompting LLMs for clarification dialogues.}
    \label{tab:cq_example}
\end{table*}

\begin{table*}[!t]
    \centering
    \begin{tabular}{p{0.97\textwidth}}
    \toprule
    \textbf{\textit{Target-guided Dialogues}}     \\
    \midrule
    \textbf{Standard Prompting}: Given the target topic and the conversation history, generate the response.\\
    \midrule
    \textbf{Proactive Prompting}: Given the target topic and the conversation history, predict the appropriate next topics that can bridge the current conversation topics to approach the target topics smoothly. Then based on the predicted next topics, generate the response. Please reply by completing the output template "The next topics are []. The response is".\\
    \midrule
    \textbf{Proactive Prompting w/ CoT}:  Given the target topic and the conversation history, consider the relationship between the current conversation topics and the target topics, and then predict the appropriate next topics that can bridge the current conversation topics to approach the target topics smoothly. Then based on the predicted next topics, generate the response. Please reply by completing the output template "The current topics are []. To bridge the current topics with the target topics, the next topics are []. Based on the predicted next topics, the response is".\\
    \midrule
    \textbf{Sample}:\\
    \makecell{\multicolumn{1}{p{0.95\textwidth}}{Target topic: "Chicken"} \\
    \multicolumn{1}{p{0.95\textwidth}}{Conversation history: ["User": "I also remodel homes when I am not out bow hunting.", "System": "That's neat. When I was in high school I placed 6th in 100m dash!", "User": "That's awesome. Do you have a favorite season or time of year?"]}}\\
    \midrule
    \textbf{Demonstration (Standard)}: I do not. But I do have a favorite meat since that is all I eat exclusively.\\
    \midrule
    \textbf{Demonstration (Proactive)}: The next topics are ["eat", "meat"]. The response is "I do not. But I do have a favorite meat since that is all I eat exclusively."\\
    \midrule
    \textbf{Demonstration (ProCoT)}: The current topics are ["season", "time", "year"]. To bridge the current topics with the target topics, the next topics are ["eat", "meat"]. Based on the predicted next topics, the response is "I do not. But I do have a favorite meat since that is all I eat exclusively."\\
    \bottomrule
    \end{tabular}
    \caption{Examples of prompting LLMs for target-guided dialogues.}
    \label{tab:tg_example}
\end{table*}

\begin{table*}[!t]
    \centering
    \begin{tabular}{p{0.97\textwidth}}
    \toprule
    \textbf{\textit{Non-collaborative Dialogues}}     \\
    \midrule
    \textbf{Standard Prompting}: Assume you are the seller. Given the item description, the target selling price, and the conversation history, generate the response.\\
    \midrule
    \textbf{Proactive Prompting}: Assume you are the seller. Given the item description, the target selling price, and the conversation history, in order to reach a better deal with the buyer, first select the most appropriate set of negotiation strategies and the most appropriate dialogue act to reach the bargain price. Based on the selected negotiation strategies and dialogue act, generate the response. The reply should be in the form "The most appropriate set of negotiation strategies is [] and the most appropriate dialogue act is []. Based on the selected negotiation strategies and dialogue act, the response is"\\
    \midrule
    \textbf{ProCoT Prompting}: Assume you are the seller. Given the item description, the target selling price, and the conversation history, in order to reach a better deal with the buyer, first analyse the current negotiation progress and consider an appropriate negotiation goal, then select the most appropriate set of negotiation strategies and the most appropriate dialogue act to reach the bargain price. Based on the selected negotiation strategies and dialogue act, generate a response. The reply should start with the analysis of the current negotiation progress and an appropriate goal, and then follow by "To reach this goal, the most appropriate set of negotiation strategies is [] and the most appropriate dialogue act is []. Based on the selected negotiation strategies and dialogue act, the response is"\\
    \midrule
    \textbf{Pre-defined Dialogue Acts}: [Answer a question, Proposing a counter price, Accept the offer, ...]\\
    \midrule
    \textbf{Pre-defined Negotiation Strategies}: [Communicate politely, Build rapport, Show dominance, ...]\\
    \midrule
    \textbf{Sample}:\\
    \makecell{
    \multicolumn{1}{p{0.95\textwidth}}{Item description: "Phone charge two devices simultaneously on the go. This vehicle charger with an additional USB port delivers enough power to charge two devices at once. The push button activated led connector light means no more fumbling in the dark trying to connect your device. Auto detect IC technology automatically detects the device type and its specific charging needs for improved compatibility. And the built in indicator light illuminates red to let you know the charger is receiving power and the power socket is working properly. Verizon car charger with dual output micro USB and led light." Target selling price: 10.} \\
    \multicolumn{1}{p{0.95\textwidth}}{Conversation history: ["Buyer": "Hi, not sure if the charger would work for my car. can you sell it to me for \$5?", "Seller": "It will work, I have never seen a car without a cigarette lighter port.", "Buyer": "Still, can i buy it for \$5? I'm on a tight budget."]}} \\    
    \midrule
    \textbf{Demonstration (Standard)}: I think the lowest I would want to go is 8.\\
    \midrule
    \textbf{Demonstration (Proactive)}: The most appropriate set of negotiation strategies is ["Propose price", "Show dominance", 'Certainty words'] and the most appropriate dialogue act is ["Proposing a counter price"]. Based on the selected negotiation strategies and dialogue act, the response is "I think the lowest I would want to go is 8."\\
    \midrule
    \textbf{Demonstration (ProCoT)}: The buyer proposes a low price, which is unacceptable. The next step should assertively raise the bargain price.  To reach this goal, the most appropriate set of negotiation strategies is ["Propose price", "Show dominance", 'Certainty words'] and the most appropriate dialogue act is ["Proposing a counter price"]. Based on the selected negotiation strategies and dialogue act, the response is "I think the lowest I would want to go is 8."\\
    \bottomrule
    \end{tabular}
    \caption{Examples of prompting LLMs for non-collaborative dialogues.}
    \label{tab:nc_example}
\end{table*}

\end{document}